\documentclass[numbered]{trbunofficial}
\usepackage{graphicx}
\usepackage{booktabs} 

\usepackage{amsmath,amsfonts}
\usepackage{amssymb}
\usepackage{algorithmic}
\usepackage{array}
\usepackage{caption}
\usepackage[font=normalsize,labelfont=sf,textfont=sf]{subfig}
\usepackage{textcomp}
\usepackage{stfloats}
\usepackage{url}
\usepackage{verbatim}
\usepackage{graphicx}
\usepackage{booktabs}
\usepackage{multirow}
\usepackage{array}

\usepackage{makecell}
\def\BibTeX{{\rm B\kern-.05em{\sc i\kern-.025em b}\kern-.08em
    T\kern-.1667em\lower.7ex\hbox{E}\kern-.125emX}}
\usepackage{balance}

\usepackage{xcolor}

\usepackage[hidelinks]{hyperref}

\AuthorHeaders{Kosieradzki and Choi}
\title{TrajFlow: A Generative Framework for Occupancy Density Estimation Using Normalizing Flows}

\author{%
  \textbf{Mitch Kosieradzki}\\
  Department of Computer Science\\
  University of Minnesota\\
  kosie011@umn.edu\\
  \hfill\break
  \hfill\break%
  \textbf{Seongjin Choi, Ph.D.}\\
  Department of Civil, Environmental, and Geo- Engineering\\
  University of Minnesota\\
  chois@umn.edu\\
}

\TotalWords{6013}
\nolinenumbers

\begin{document}
\maketitle

\section{Abstract}

For intelligent transportation systems and autonomous vehicles to operate safely and efficiently, they must reliably predict the future motion and trajectory of surrounding agents within complex traffic environments.
%
%
At the same time, the motion of these agents is inherently uncertain, making accurate prediction difficult.
In this paper, we propose \textbf{TrajFlow}, a generative framework for estimating the occupancy density of dynamic agents.
%
Our framework utilizes a causal encoder to extract semantically meaningful embeddings of the observed trajectory, as well as a normalizing flow to decode these embeddings and determine the most likely future location of an agent at some time point in the future. 
Our formulation differs from existing approaches because we model the marginal distribution of spatial locations instead of the joint distribution of unobserved trajectories. The advantages of a marginal formulation are numerous. First, we demonstrate that the marginal formulation produces higher accuracy on challenging trajectory forecasting benchmarks. Second, the marginal formulation allows for fully continuous sampling of future locations. Finally, marginal densities are better suited for downstream tasks as they allow for the computation of per-agent motion trajectories and occupancy grids, the two most commonly used representations for motion forecasting. We present a novel architecture based entirely on neural differential equations as an implementation of this framework and provide ablations to demonstrate the advantages of a continuous implementation over a more traditional discrete neural network based approach. The code is available at \url{https://github.com/UMN-Choi-Lab/TrajFlow}. 
\hfill\break%
\noindent\textit{Keywords}: Motion Forecasting, Intelligent Transportation Systems, Autonomous Vehicles, Neural Differential Equations, Normalizing Flows
\newpage

\section{Introduction}
For intelligent transportation systems (ITS) and autonomous vehicles (AVs) to operate safely and efficiently, they must reliably predict the future motion and trajectory of surrounding agents within complex traffic environments.
Accurate trajectory prediction is a fundamental ITS problem, which plays a key role in enabling a wide range of ITS subsystems aimed at improving mobility, safety, and efficiency in transportation networks \cite{tyagi2022introduction}. For instance, accurate trajectory forecasting enhances Advanced Driver Assistance Systems (ADAS) by enabling functionalities such as lane-keeping, adaptive cruise control, and proactive collision avoidance through the anticipation of nearby vehicles' behavior. Within Advanced Traffic Management Systems (ATMS), such predictions enhance vehicle-level control strategies at intersections and during congestion, where precise interactions between agents must be coordinated. Moreover, as ITS expands to include connected infrastructure, predictive models at the microscopic level will also improve real-time routing, intersection control, and pedestrian safety measures.


%
However, predicting future agent trajectories remains challenging due to inherent uncertainties and the stochastic nature of human and vehicular movements in transportation networks.
Early attempts at trajectory prediction, such as Social LSTM \cite{alahi2016social}, SR-LSTM \cite{zhang2019sr}, and Social Attention \cite{vemula2018social}, focused on \textbf{deterministically forecasting} a single future trajectory for each participant. These deterministic regressors struggled to capture the inherent uncertainty in future motion prediction.
To address this limitation, research has shifted towards \textbf{probabilistic forecasting}. For example, Pajouheshgar et al. \cite{pajouheshgar2018back} utilized spatio-temporal convolutional neural networks (CNNs) \cite{ji20133d} to predict multiple future trajectories probabilistically. Trajectron++ \cite{salzmann2020trajectron} combined spatio-temporal CNNs with recurrent neural networks (RNNs) for multi-modal forecasting, while FloMo \cite{scholler2021flomo} employed RNNs and normalizing flows \cite{tabak2013family} for trajectory density estimation. 

Normalizing flows have emerged as a promising probabilistic framework due to their ability to compute exact likelihoods through sequence of invertible transformations. Unlike other probabilistic models that rely on predefined distributions (e.g., Gaussian Mixtures \cite{6232277}) or other generative models like Variational Auto-Encoders (VAEs) \cite{kingma2013auto} and Generative Adversarial Networks (GANs) \cite{goodfellow2020generative} that only rely on approximate inference or implicit density modeling, normalizing flows allow for direct likelihood estimation with efficient sampling. This property makes them particularly well-suited for safety-critical applications. In such a context, the ability to explicitly model probability distribution enables robust decision-making under quantifiable uncertainty, enhances risk assessment, and supports scenario planning for rare or dangerous events.


\begin{figure}[tb]
    \centering
    \includegraphics[width=0.7\textwidth]{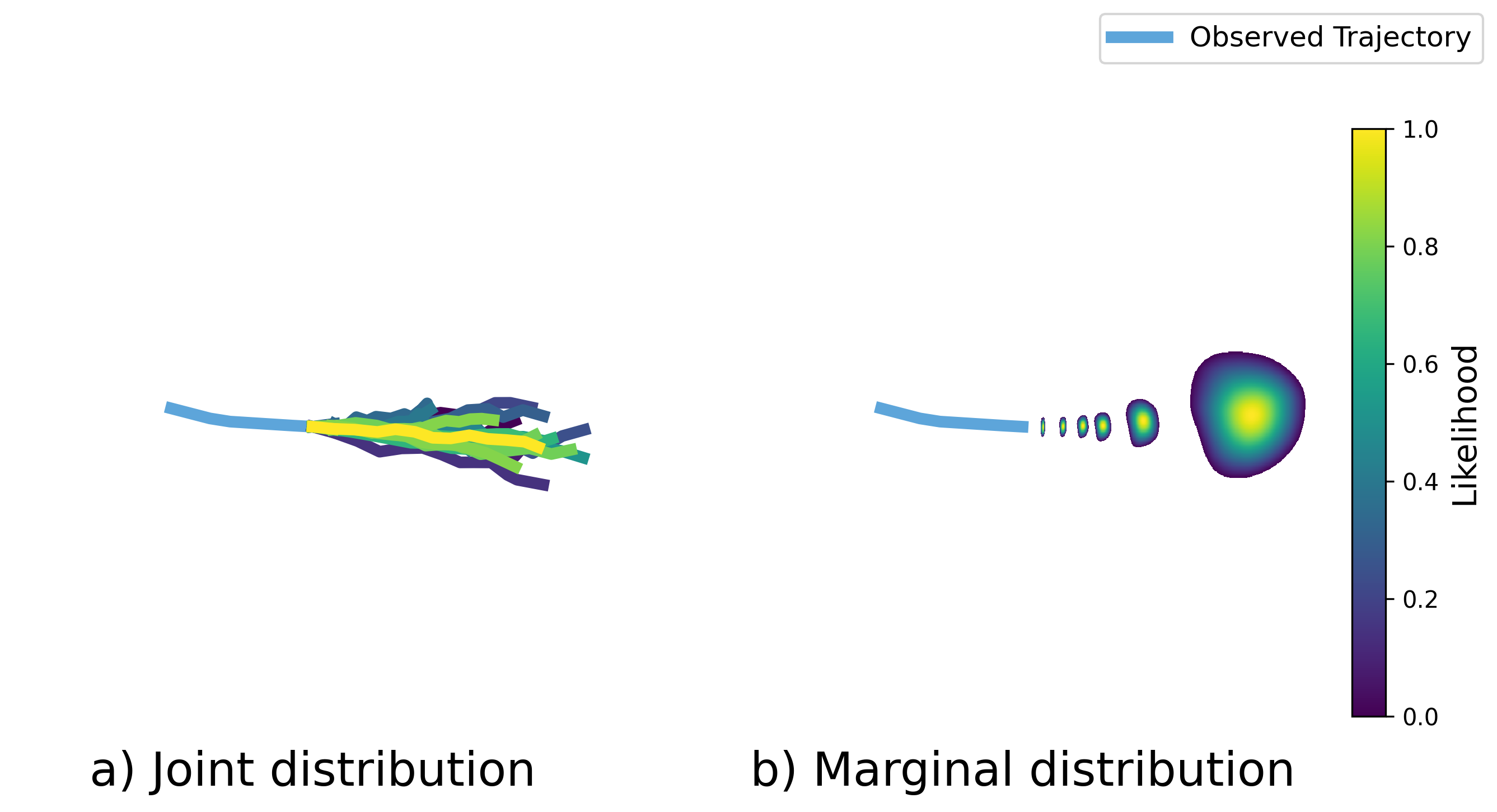}
    \caption{Trajectory forecasts of the \textbf{TrajFlow} framework when the model is trained to learn the joint distribution (left) and marginal distribution (right). 
    }
    \label{fig:forecasts}
\end{figure}

Yet, while probabilistic forecasting, including normalizing flows, has made significant progress, important challenges remain in how uncertainty is modeled and leveraged in practice.
Many existing approaches primarily focus on modeling the joint density of these locations (as shown in Figure \ref{fig:forecasts} (a)) and often operate under a fixed and discrete forecast horizon. Although the joint formulation facilitates the sampling of smooth future motion trajectories and maintains diversity in the samples, 
obtaining marginal spatial distributions from the joint density is computationally intractable.
%
In contrast, marginalized spatial densities (as shown in Figure \ref{fig:forecasts} (b)) 
allow for an infinite and continuous forecast horizon, offering significant benefits for downstream tasks by supporting both trajectory reasoning and occupancy-based representations \cite{thrun1996integrating}.
%
Integrating these representations allows systems to maintain comprehensive situational awareness, which is essential for solving complex planning and control challenges.

For example, in autonomous vehicles, these representations help intelligent agents plan safe, collision-free motion by predicting the future motion and occupancy of surrounding vehicles. Similarly, in transportation networks, traffic accidents are typically concentrated in areas known as black-spots \cite{ghadi2019comparative, ghadi2017comparison}. Previous work has explored the use of the Poisson-Tweedie model \cite{debrabant2018identifying}, screening \cite{sandhyavitri2017three}, clustering \cite{sandhyavitri2017three}, and crash prediction \cite{sandhyavitri2017three} as black-spot identification methods. However, no standardized method for black-spot identification has yet been proposed that applies to all road types \cite{ghadi2017comparison}. Marginal spatial density estimations facilitate the computation of collision probabilities, enabling intelligent transportation systems to detect black-spots for a wide variety of road types automatically.


Additionally, many existing methods predominantly use discrete neural networks in the construction of their forecasting models. However, trajectory data is inherently continuous, and discrete modeling approaches often fail to fully capture all of the information present in such data. One way to address this limitation is by leveraging neural differential equations \cite{chen2018neural, kidger2020neural} in the model's construction. Neural differential equations are a continuous modeling paradigm that aligns with the continuous nature of trajectory data.

\begin{figure*}[!t]
    \centering
    \includegraphics[width=1.0\textwidth]{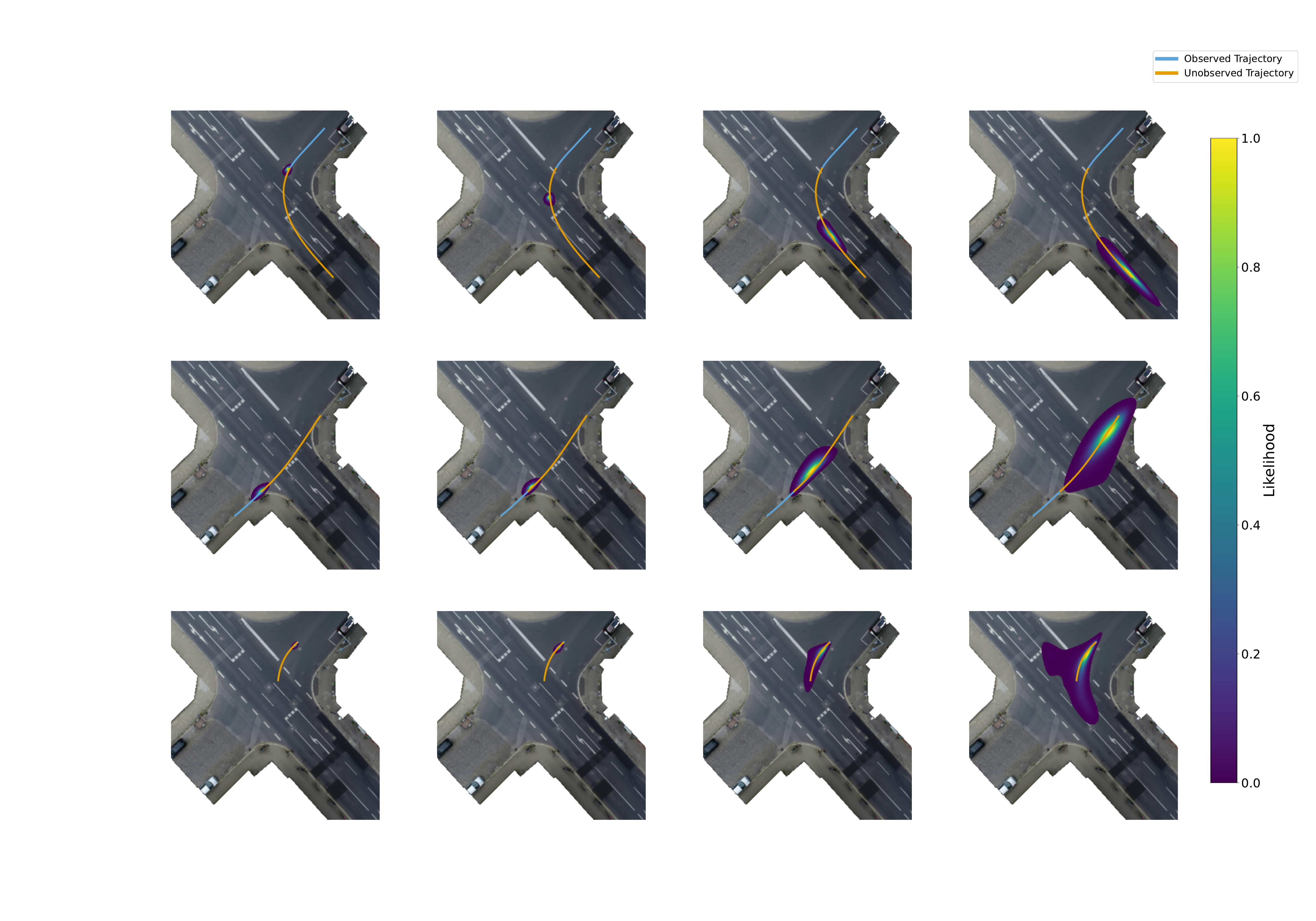}
    \caption{Our framework learns the change in occupancy density over time conditioned on an observed trajectory. Here we see the evolution of occupancy densities after $0$, $33$, $66$ and $99$ time steps into the future.} 
    \label{fig:spatial-density-evolution}
\end{figure*}

In this work, we introduce \textbf{TrajFlow}, a novel framework for the estimation of probabilistic occupancy densities using normalizing flows (see Figure \ref{fig:spatial-density-evolution} for examples). Our framework directly models the marginal occupancy density for each future agent location, rather than the joint distribution over future motion trajectories. 
\textbf{TrajFlow} employs a normalizing flow conditioned on an embedding produced by a causal encoder to model these densities. We explore both discrete and continuous implementations of the \textbf{TrajFlow} framework and provide ablation studies to demonstrate the performance advantages of the continuous approach. By combining neural differential equations with a marginal density formulation, \textbf{TrajFlow} enables a fully continuous framework, allowing it to encode continuous trajectories and sample future occupancy densities seamlessly over continuous time. Inference-time algorithms empower \textbf{TrajFlow} to generate probabilistic occupancy maps and infer likely future motion trajectories in a principled and tractable manner.

\subsection{Summary of Our Contributions}

The novelty of this paper can be summarized as follows:

\begin{enumerate}[leftmargin=*]
     \item We propose a novel density formulation for \textbf{TrajFlow}. Specifically, while previous works modeled the joint distribution of unobserved trajectories, \textbf{TrajFlow} focuses on modeling the marginal distribution of spatial locations.
    \item We present a novel architecture based entirely on neural differential equations as an implementation of the \textbf{TrajFlow} framework, providing continuous processing capabilities on the observed trajectory,
    \item We conduct an ablation study to compare the performance of a neural controlled differential equation with a recurrent neural network for the causal encoder and to evaluate the differences between using a neural ordinary differential equation as a continuous normalizing flow versus a discrete one.
    \item The combination of neural differential equations and marginal density modeling facilitates a fully continuous forecasting framework and enables \textbf{TrajFlow} to achieve state-of-the-art performance on challenging trajectory forecasting datasets, outperforming existing models.
    \item We demonstrate how inference-time algorithms, such as additive fusion and top-k sampling, can be leveraged by \textbf{TrajFlow} to construct occupancy grids and smooth motion trajectories, the two most common representations for motion forecasting.
\end{enumerate}

\section{Background}

\subsection{Deep Neural Networks and Deep Generative Models}
In recent years neural networks have emerged as the dominant modeling paradigm for a wide variety of tasks. For example, convolutional neural networks (CNNs) \cite{LeCun1995} have seen immense success in solving various computer vision tasks such as object classification \cite{Krizhevsky2012,he2016deep,huang2017densely}, object detection \cite{girshick2014rich,redmon2016you}, and semantic segmentation \cite{long2015fully,ronneberger2015u}. Recurrent neural networks (RNNs), such as those with gated recurrent units (GRU) \cite{cho2014learning}, have become a popular choice for modeling tasks involving sequential data found in natural language processing \cite{sutskever2014sequence,graves2013generating} and time series forecasting \cite{salinas2020deepar}.

Neural networks were initially applied to trajectory forecasting as deterministic regressors \cite{alahi2016social, zhang2019sr, vemula2018social}, tasked with predicting the single most likely future trajectory based on observed inputs. While these methods may be effective in structured environments, they exhibit significant limitations in more complex and dynamic scenarios. Specifically, deterministic models struggle to account for the inherent uncertain nature of real-world motion, where multiple plausible future trajectories can emerge due to stochastic factors and complex interactions between agents and their environments \cite{gupta2018social, salzmann2020trajectron, scholler2021flomo, socialvae2022}.

The emergence of generative modeling techniques has marked a transformative shift in trajectory forecasting research \cite{choi2024gentle}. Unlike deterministic approaches, generative models are designed to learn and represent the underlying probability distribution of future trajectories, rather than focusing on a single prediction. Generative models have enabled a comprehensive understanding of uncertainty and multimodality in future motion, providing researchers with a powerful framework for simulating and analyzing complex motion patterns. This capability has made generative approaches not only more flexible and robust but also indispensable for addressing challenges that traditional deterministic methods could not overcome.

Over the years, many different neural architectures have been proposed for generative modeling. For instance, generative adversarial networks (GANs) \cite{goodfellow2020generative} utilize a generative network that learns to transform a sample from a latent, often Gaussian, distribution into a sample from the data distribution. To ensure the generative network correctly models the data distribution, a discriminative network is employed to distinguish between real samples and those generated by the generative network, improving the quality of the generated samples through adversarial training. Variational auto-encoders (VAEs) \cite{kingma2013auto} train an encoder to map data points to a latent space and a decoder to reconstruct data from samples drawn from this latent space. Indeed, many latent space methods have been proposed, but they often can only sample data points from the learned distribution, lacking the ability to compute exact likelihoods.

\subsection{Normalizing Flows}
On the other hand, Normalizing Flows \cite{tabak2013family}, or flow-based generative models, are a latent space method that not only allows for the sampling of new data points but also allows for exact likelihood computation. They achieve this by directly modeling the mapping
\begin{equation}
z = f(x)
\end{equation}
that maps samples from the data distribution, $x \sim p_{x}(x)$, to samples from the latent distribution $z ~\sim p_{z}(z)$. By defining the map $f$ such that it is differentiable, the density $p_{x}$ can be transformed into the density $p_{z}$ by a change of variables \cite{Rudin2006,Bogachev2007}
\begin{equation} 
\begin{split} 
    &p_{z}(z) = p_{x}(x)|\det J_{f}(x)|^{-1}\\
    &\log(p_{z}(z)) = \log(p_{x}(x)) - \log(|\det J_{f}(x)|)    
\end{split}
\end{equation}
where $J_{f}$ is the Jacobian of the map $f$. Note if $f$ is invertible then it is also possible to obtain $p_{x}$ from $p_{z}$ in a similar manner
\begin{equation}
    p_{x}(x) = p_{z}(z)|\det J_{f^{-1}}(z)|^{-1}
\end{equation}
Computing the determinant of the Jacobian matrix generally has $\mathcal{O}(n^3)$ time complexity, however, if we design $f$ such that $J_{f}$ is an upper or lower triangular matrix, we can efficiently compute the determinant with $\mathcal{O}(n)$ time complexity.

\subsection{Neural Differential Equations}

Neural networks such as residual networks transform a hidden state $h_{t}$ along a manifold through a sequence of discrete transformations described by
\begin{equation}
    h_{t+1} = h_{t} + g_{\theta_t}(h_{t}).
\end{equation}
This sequence of discrete transformations is an Euler discretization of a continuous transformation \cite{lu2018beyond}\cite{haber2017stable}\cite{ruthotto2020deep}. As $t \to \infty$ we can specify the dynamics of this continuous transformation as an ordinary differential equation parameterized by a neural network \cite{chen2018neural}
\begin{equation}
    \frac{d h(t)}{dt} = g_{\theta}(h(t), t).
\end{equation}
Given a network input, $h(t_{0})$, we can use a numerical integrator to solve the initial value problem
\begin{equation}
    h(t_{1}) = h(t_{0}) + \int_{t_0}^{t_1} g_{\theta}(h(t), t) \, dt
\end{equation}
to obtain the network output $h(t_{1})$.

Instead of applying reverse mode auto-differentiation through the operations of the numerical integrator to obtain the gradients of the loss function $\mathcal{L}$ with respect to $\theta$, these gradients can also be computed by solving another ordinary differential equation initial value problem \cite{pontryagin1962mathematical}:
\begin{equation}
    \frac{d \mathcal{L}}{d\theta} = -\int_{t_1}^{t_0} \left(\frac{d \mathcal{L}}{dh(t)}\right)^T \frac{d g_{\theta}(h(t), t)}{d\theta} \, dt,
\end{equation}
where $\frac{d \mathcal{L}}{dh(t)}$ is the adjoint state of the differential equation and the initial value is $\frac{d \mathcal{L}}{dh(t_1)}$. This method allows the gradients of the neural differential equation to be computed efficiently with $\mathcal{O}(1)$ memory complexity.

\subsection{Neural Differential Equations for Continuous Normalizing Flows}

One interesting application of neural ordinary differential equations is continuous normalizing flows \cite{grathwohl2018ffjord}. Instead of using a neural network for the map $f$, as in discrete normalizing flows, this map is specified by a neural ordinary differential equation. The density $p_{x}$ can be transformed into the density $p_{z}$ by an instantaneous change of variables \cite{chen2018neural}
\begin{equation}
    \log(p(h(t_{1}))) = \log(p(h(t_{0})) - \int_{t_0}^{t_1}\operatorname{Tr}\left( J_{f}(h(t))\right) \, dt
\end{equation}
where $z = h(t_{1})$ and $x = h(t_{0})$. In this way, the neural ordinary differential equation defines the dynamics of a smooth flow of probability mass from the data distribution $p_{x}(x)$ to the latent distribution, $p_{z}(z)$. While computing the trace of the Jacobian can usually be done with $\mathcal{O}(n^2)$ time complexity, utilizing a Hutchinson estimator \cite{hutchinson1989stochastic} can reduce the time complexity to $\mathcal{O}(n)$ without the need to restrict the form that the Jacobian matrix must take.

If neural ordinary differential equations can be seen as a continuous analog of residual networks, then neural controlled differential equations can be seen as a continuous analog of recurrent neural networks \cite{kidger2020neural}. Given $\mathcal{C}:[t_0,t_1] \to \mathbb{R}^v$, a continuous function of bounded variation where $t_0, t_1 \in \mathbb{R}$ with $t_0 < t_1$ and the continuous function $\xi_{\theta}:\mathbb{R}^w \to \mathbb{R}^{w \times v}$, we can define a neural controlled differential equation as
\begin{equation}
    h(t_1) = h(t_0) + \int_{t_0}^{t_1} \xi_{\theta}(h(t))d\mathcal{C}_t
\end{equation}
Here the integral is a Riemann–Stieltjes integral and $\xi_{\theta}(h(t))d\mathcal{C}_t$ is the vector field defining the dynamics of the transformation of the hidden state $h(t)$ along the manifold. The Riemann-Stieltjes integral can be related to the Riemann integral with the following identity \cite{Rudin2006}
\begin{equation}
    \int_{t_0}^{t_1} \xi_{\theta}(h(t))d\mathcal{C}_t = \int_{t_0}^{t_1} \xi_{\theta}(h(t))\mathcal{C}'(t) \, dt
\end{equation}

\section{Methodology}

\subsection{Problem Statement}

The observed motion of a dynamic agent can be defined as the finite sequence $O = (o^0,...,o^T)$ of observed spatial positions $o^t = (x^t,y^t)$ across discrete timesteps $t \in \{0,...,T\}$. Likewise, the unobserved motion of a dynamic agent is the finite sequence $U = (u^1,...,u^S)$ of unobserved positions $u^s = (x^{T+s},y^{T+s})$ across discrete timesteps $s \in \{1,...,S\}$. We can forecast the unobserved future position of a dynamic agent at time $s$ by modeling the conditional likelihood $p(u^s|O)$ and sampling to obtain $u^s ~\sim p(u^s|O)$. If we assume conditional independence among the unobserved future positions then we can compute the likelihood of an unobserved motion sequence through the density:
\begin{equation}
P(U|O) = \prod_{s=1}^{S} p(u^s|O).
\end{equation}

\subsection{TrajFlow}
\begin{figure*}[t]
    \centering
    \includegraphics[width=\textwidth]{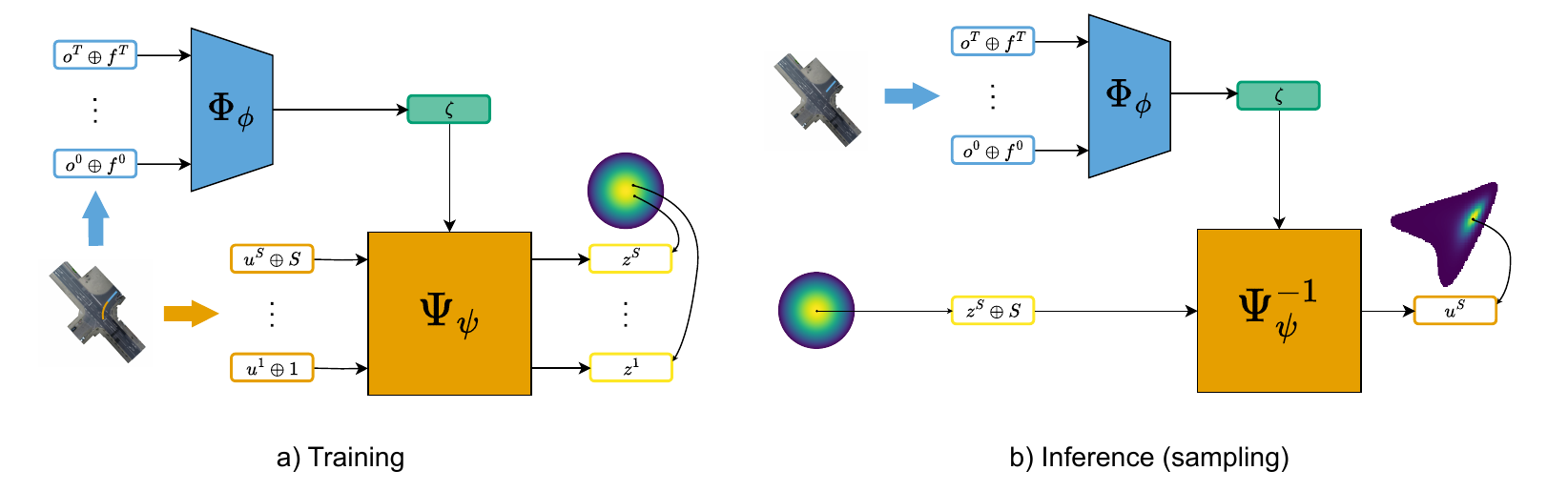}
    \caption{\textbf{TrajFlow} architecture. The causal encoder $\Phi_{\phi}$ encodes the observed trajectory $O$ and features $F$ into the embedding $\zeta$. At training time, the unobserved trajectory $U$ is passed as input into the flow $\Psi_{\psi}$ along with the times $s$ associated with each unobserved position in the trajectory and the embedding $\zeta$. The model is trained to maximize the likelihood of the unobserved positions. At inference time, samples from the latent distribution along with the times $s$ and embedding $\zeta$ are passed as input into the inverse flow $\Psi^{-1}_{\psi}$ to produce synthetic samples from the data distribution.}
    \label{fig:train-architecture}
\end{figure*}

In this section, we present \textbf{TrajFlow}, our generative framework for occupancy density estimation. The objective of our framework is to learn $p(u^{s}_i|O_i,F_i)$, the likelihood of the agent $A_i$ being at location $u^{s}_i$ at time $s$ given the observed trajectory $O_i$ and features $F_i$. The feature sequence $F_i=(f_i^0,...,f_i^T)$ contains the feature vectors $f_i^t = (\dot{x}_i^t, \dot{y}_i^t, \ddot{x}_i^t, \ddot{y}_i^t, \theta_i^t)$ for $t \in \{0,...,T\}$. Here $\dot{x}_i^t$ and $\dot{y}_i^t$ are the $x$ and $y$ velocity of agent $A_i$ at time $t$, $\ddot{x}_i^t$ and $\ddot{y}_i^t$ are the $x$ and $y$ acceleration of agent $A_i$ at time $t$, and $\theta_i^t$ is the heading of agent $A_i$ at time $t$. An overview of our architecture is given in Figure \ref{fig:train-architecture}.

We learn the likelihood $p(u^{s}_i|O_i,F_i)$ by training two models: a \textbf{causal encoder} $\Phi$ parameterized by $\phi$ that encodes the observed trajectory $O_i$ and features $F_i$ as a semantically meaningful embedding $\zeta$, and a \textbf{normalizing flow} $\Psi$ parameterized by $\psi$ that allows us to sample $p(u^{s}_i|O_i,F_i)$ given a sample from a latent distribution $z^s ~\sim p_{z}(z)$. Our framework can be summarized by the following equations:
\begin{equation}
\begin{split}
    \zeta &= \Phi_{\phi}(O_i, F_i) \\
    u^s_i &= \Psi_{\psi}^{-1}(\zeta, s, z^s)
\end{split}
\end{equation}
Both the causal encoder $\Phi_{\phi}$ and the normalizing flow $\Psi_{\psi}$ can be parameterized as a discrete neural network or a continuous neural differential equation.

\subsection{Discrete Model}
\subsubsection{Causal Encoder}

In the discrete parameterization, we used GRU \cite{cho2014learning} as our causal encoder. GRU is defined by the following set of equations:
\begin{equation}
\begin{split}
    r_t &= \sigma(W_{ir} x_t + b_{ir} + W_{hr} h_{t-1} + b_{hr}) \\
    z_t &= \sigma(W_{iz} x_t + b_{iz} + W_{hz} h_{t-1} + b_{hz}) \\
    \tilde{h}_t &= \tanh(W_{in} x_t + b_{in} + r_t \odot (W_{hn} h_{t-1} + b_{hn})) \\
    h_t &= (1 - z_t) \odot \tilde{h}_t + z_t \odot h_{t-1}
\end{split}
\end{equation}
$h_t$ is the hidden state at time $t$ and $x_t$ is the input at time $t$. Note that $h_0 = 0$. $\tilde{h}_t$ is the candidate hidden state given the current information $x_t$. $r_t$ is the reset gate and is responsible for determining how much information from the previous hidden state $h_{t-1}$ should be forgotten. $z_t$ is the update gate and is responsible for determining how much information from the previous hidden state $h_{t-1}$ and the candidate hidden state $\tilde{h}_t$ should be incorporated into the current hidden state $h_{t}$.

\subsubsection{Normalizing Flow}

We used a stack of affine coupling layers \cite{dinh2016density, dinh2014nice} for our normalizing flow in the discrete parameterization. The forward pass of an affine coupling layer can be formulated as follows:
\begin{equation}
\begin{split}
    y_{1:d} &= x_{1:d} \\
    y_{d+1:D} &= x_{d+1:D} \odot \exp(s_\theta(x_{1:d})) + t_\theta(x_{1:d}).
\end{split}
\end{equation}
The inverse pass is formulated as follows:
\begin{equation}
\begin{split}
    x_{1:d} &= y_{1:d} \\
    x_{d+1:D} &= (y_{d+1:D} - t_\theta(y_{1:d})) \odot \exp(-s_\theta(y_{1:d})).
\end{split}
\end{equation}
Here $x_{1:D}$ is the layer input and $y_{1:D}$ is the layer output. $t_\theta$ and $s_\theta$ are simple multi-layer perceptrons (MLP). The Jacobian matrix of the affine coupling layer is formulated as:
\begin{equation}
    \frac{d y}{d x^T} = 
\begin{bmatrix}
\mathbb{I}_d & 0 \\
\frac{d y}{d x^T} & \operatorname{diag}(\exp[s_\theta(x_{1:d})])
\end{bmatrix}.
\end{equation}
Since this matrix is lower triangular, we can efficiently compute its determinant with $\mathcal{O}(n)$ time complexity using the following equation:
\begin{equation}
\begin{split}
\left|\det\left(\frac{d y}{d x^T}\right)\right| &= \prod_{i=1}^{D}\left[\frac{d y}{d x^T}\right]_{i,i}=\prod_{i=1}^{d}\exp[s_\theta(x_{1:d})]_{i,i} \\
&= \exp\left(\sum_{i=1}^{d} [s_\theta(x_{1:d})]_{i,i}\right)
\end{split}.
\end{equation}
After each affine coupling layer, we applied a moving average batch norm \cite{grathwohl2018ffjord,dinh2016density} to the activations. This form of batch normalization learns to rescale each dimension and has the effect of stabilizing training when training a conditional distribution. Moving average batch norm is defined as
\begin{equation}
\begin{split}
    &\text{MBN}(x) = \frac{x - \mu}{\sigma} \gamma + \beta \\
    &\text{MBN}^{-1}(y) = \frac{y - \beta}{\gamma} \sigma + \mu
\end{split}
\end{equation}
The main difference between moving average batch norm and standard batch norm \cite{ioffe2015batch} is that $\mu$ and $\sigma$ are running averages of the batch mean and standard deviation. $\gamma$ and $\beta$ are trainable model parameters. The log determinant of moving average batch norm can be defined as
\begin{equation}
\log \left( \left|\det\frac{\partial \text{MBN}(x)}{\partial x} \right| \right) = \sum_{i} \log |\gamma_i| - \log |\sigma_i|
\end{equation}

\subsection{Continuous Model}

\subsubsection{Causal Encoder}

We used a neural controlled differential equation (CDE) \cite{kidger2020neural} as our causal encoder in the continuous parameterization. The function $\xi_{\theta}$ was implemented as a simple multi-layer perceptron. The initial hiden state, $h(t_0)$, was produced from $o^0$ and $f^0$ through the embedding matrix $W_{embed}$. We used a natural cubic spline interpolated through the observed trajectory $O$ and features $F$ as the control signal $\mathcal{C}$. The $i$-th piece of the spline can be expressed as \cite{Bartels1998}:
\begin{equation}
    \mathcal{C}_i(t) = \alpha_i + \beta_it + \gamma_it^2 + \delta_it^3,
\end{equation}
where $t \in [0, 1]$ and $i \in \{0, \ldots, n-1\}$. The derivative of the $i$-th piece of the spline with respect to $t$ can be expressed as:
\begin{equation}
    \mathcal{C}_i'(t) = \beta_i + 2\gamma_it + 3\delta_it^2.
\end{equation}
Notice that
\begin{equation}
\begin{split}
    \mathcal{C}_i(0) &= c_i = \alpha_i \\
    \mathcal{C}_i(1) &= c_{i+1} = \alpha_i + \beta_i + \gamma_i + \delta_i \\
    \mathcal{C}'_i(0) &= \mathcal{D}_i = \beta_i \\
    \mathcal{C}'_i(1) &= \mathcal{D}_{i+1} = \beta_i + 2\gamma_i + 3\delta_i,
\end{split}
\end{equation}
and so we can express $\alpha_i$, $\beta_i$, $\gamma_i$ and $\delta_i$ as
\begin{equation}
\begin{split}
    \alpha_i &= c_i \\
    \beta_i &= \mathcal{D}_i \\
    \gamma_i &= 3(c_{i+1} - c_i) - 2\mathcal{D}_i - \mathcal{D}_{i + 1} \\
    \delta_i &= 2(c_i - c_{i+1}) + \mathcal{D}_i + \mathcal{D}_{i+1}.
\end{split}
\end{equation}
Let the second derivatives be equal at the points
\begin{equation}
    \mathcal{C}''_i(0) = \mathcal{C}''_{i-1}(1),
\end{equation}
and equal to zero at the endpoints
\begin{equation}
\begin{split}
    \mathcal{C}''_0(0) &= 0 \\
    \mathcal{C}''_{n-1}(1) &= 0.
\end{split}
\end{equation}
The spline coefficients can then be obtained by solving the following tridiagonal linear system \cite{Bartels1998}:
\begin{equation}
\small
\begin{bmatrix}
2 & 1 & 0 & \cdots & 0 & 0 \\
1 & 4 & 1 & \cdots & 0 & 0 \\
0 & 1 & 4 & \cdots & 0 & 0 \\
\vdots & \vdots & \vdots & \ddots & \vdots & \vdots \\
0 & 0 & 0 & \cdots & 4 & 1 \\
0 & 0 & 0 & \cdots & 1 & 2
\end{bmatrix}
\begin{bmatrix}
\mathcal{D}_0 \\
\mathcal{D}_1 \\
\mathcal{D}_2 \\
\vdots \\
\mathcal{D}_{n-1} \\
\mathcal{D}_n
\end{bmatrix}
=
\begin{bmatrix}
3(c_1 - c_0) \\
3(c_2 - c_1) \\
3(c_3 - c_2) \\
\vdots \\
3(c_{n-1} - c_{n-2}) \\
3(c_n - c_{n-1})
\end{bmatrix}.
\end{equation}

\subsubsection{Normalizing Flow}

We used a neural ordinary differential equation (ODE) for our normalizing flow in the continuous parameterization \cite{chen2018neural} \cite{grathwohl2018ffjord}. The vector field $f_\theta$ was implemented as a feed-forward neural network. We used the \textbf{f}eature w\textbf{i}se \textbf{l}inear \textbf{m}odulatioin (\textbf{FiLM}) layer \cite{perez2018film} as the layers in this feed-forward network in order to condition $f_\theta$ on the embedding $\zeta$. The concat and squash layer is defined as:
\begin{equation}
\begin{split}
    \text{FiLM}(x, \zeta, t, s) &= (W_x x + b_x) \sigma(W_t t + W_s s + W_c \zeta + b_t) \\
    &\quad + (W_{bt} t + W_{bs} s + W_{bc} \zeta + b_{bt}).
\end{split}
\end{equation}
In order to learn the scale of each dimension and stabilize training, we used moving average batch norm before and after the continuous normalizing flow \cite{grathwohl2018ffjord}.

\subsection{Loss Function}

Since normalizing flows allow for exact likelihood computation, we train \textbf{TrajFlow} to maximum training sample likelihood by minimizing the negative log likelihood
\begin{equation}
    \mathcal{L} = -\frac{1}{N \cdot S} \sum_{i=1}^{N} \sum_{s=1}^{S} \log(p(u^s_i|O_i,F_i)).
\end{equation}

\section{Experiments}

We evaluate our framework with the publicly available ETH \cite{pellegrini2009you}, UTY \cite{lerner2007crowds}, and 
intersection drone (inD) \cite{bock2020ind} datasets. All datasets are curated from real-world video recordings and contain complex motion trajectories. 

The ETH and UCY datasets focus on \textbf{pedestrian trajectories} taken from recordings in city centers and university campuses. Together, these datasets cover five distinct scenes and four unique locations (ETH, Hotel, Univ, Zara 1\&2). They contain over $1,950$ individual pedestrian trajectories. This dataset is a challenging benchmark due to the social nature of pedestrian trajectories.

The inD dataset consists of high-accuracy \textbf{vehicle trajectories} extracted from high-resolution drone footage. This dataset covers four unique intersections and contains over $11,500$ intersection user trajectories. These intersection trajectories present a challenging learning problem for our framework due to the complexity and variety of traffic scenarios.

\subsection{Experiment Design}
We conducted two experiments to evaluate our model's performance on trajectory data for both pedestrians and vehicles. For the pedestrian experiment, we utilized the ETH and UCY datasets, while the vehicle experiments were conducted using the inD dataset. 

For each experiment, we trained the model with the Adam \cite{kingma2014adam} optimizer. We used a learning rate of $1 \times 10^{-3}$ and decay with a gamma of $0.999$. For models involving neural differential equations, we used the Dormand-Prince 5 numerical integrator \cite{dormand1980runge}. We used an absolute and relative tolerance of $1 \times 10^{-5}$. All experiments were carried out on a single NVIDIA RTX 6000 GPU with 48GB of GDDR6 VRAM and a 48-core 96-thread AMD EPYC Milan 7643 CPU.

\subsection{Baseline Models}
We compared our framework with various state-of-the-art models for both experiments. The baseline models include two deterministic models and four generative models:
\begin{itemize}[leftmargin=1em]
    \item \textbf{Deterministic Models}:
    \begin{itemize}[leftmargin=*]
        \item \textbf{TUTR} \cite{Shi_2023_ICCV}\footnote{\url{https://github.com/Carrotsniper/-ICCV-2023-TUTR-}} a transformer based encoder decoder architecture. The encoder encodes the relationship between motion modes while the decoder attends to the social interactions among neihbooring trajectories. Two prediction heads produce the trajectory estimations and corresponding probabilities.
        \item \textbf{PPT} \cite{lin2024progressive} \footnote{\url{https://github.com/iSEE-Laboratory/PPT}} a transformer architecture that utilizes a novel progressive pretext task (PPT) learning curriculum to progressively improve the predictive capacity of the model. The training curriculum consists of learning short-term dynamics, long-term dependencies, and full temporal patterns.
    \end{itemize}
    \item \textbf{Generative Models}:
    \begin{itemize}[leftmargin=*]
        \item \textbf{S-GAN} \cite{gupta2018social}\footnote{\url{https://github.com/agrimgupta92/sgan}} a GAN architecture based on LSTMs \cite{graves2012long} and a novel social coupling layer.
        \item \textbf{Trajectron++} \cite{salzmann2020trajectron}\footnote{\url{https://github.com/StanfordASL/Trajectron-plus-plus}} a VAE architecture that combines CNNs with RNNs for multimodal forecasting.
        \item \textbf{FloMo} \cite{scholler2021flomo}\footnote{\url{https://github.com/cschoeller/flomo_motion_prediction}} a discrete normalizing flow conditioned on embeddings produced by an RNN. This model is the most similar to discrete implementation \textbf{TrajFlow} with the main difference being the flow is implemented with monotonic rational-quadratic splines \cite{durkan2019neural} instead of affine coupling layers.
        \item \textbf{S-VAE} \cite{socialvae2022}\footnote{\url{https://github.com/xupei0610/SocialVAE}} a timewise VAE that utilizes a stochastic RNN and a social attention mechanism to condition the prediction.
    \end{itemize}
\end{itemize}

In addition, we conducted a component-level ablation study on the causal encoder and normalizing flow to assess the impact of the continuous implementation of our framework. 
We evaluated the following for configurations of our framework \textbf{GRU-DNF}, \textbf{GRU-CNF}, \textbf{CDE-DNF}, and \textbf{CDE-CNF}. \textbf{GRU-DNF} is a fully discrete implementation while \textbf{CDE-CNF} is fully continuous. \textbf{GRU-CNF} and \textbf{CDE-DNF} are both partial continuous implementations, with the former having a continuous flow and the latter having a continuous encoder.
Our findings revealed that the fully continuous implementation (\textbf{CDE-CNF}) performed better across both experiments. Consequently, we compared this implementation under both the marginal and joint density formulations.

\subsection{Pedestrian Experiment}

For the ETH/UCY datasets, we followed the common evaluation strategy \cite{gupta2018social, salzmann2020trajectron, scholler2021flomo, socialvae2022, Shi_2023_ICCV, lin2024progressive}. We sliced each trajectory into sequences of length $20$ with a step size of $1$. $8$ of the timesteps corresponded to the observed trajectory, and the remaining $12$ corresponded to the unobserved trajectory. For training, we consider trajectories that contain all $20$ time steps. However, during evaluation, we only require that a trajectory contains $10$ time steps. That is, there must be at least two future locations to predict. We evaluate using the standard leave-one-out methodology. In this methodology, the model is trained using four of the scenes and evaluated on the remaining scene. For example, the model might be trained on Hotel, Univ, Zara 1, and Zara 2 and evaluated on ETH.

As in \cite{chang2019argoverse} we found it necessary to provide rotational invariance by rotating trajectories around $o^T$ such that the last relative displacement, $o^T - o^{T-1}$, is aligned with the unit vector $(1,0)$. For the joint formulation only, we also found it necessary to provide translational invariance by learning the distribution over the relative displacement instead of the sequence of spatial locations. After sampling we rotate the predicted trajectories back and if necessary convert the relative displacements back to a sequence of spatial locations.

To enhance the diversity of our training data, we apply random scaling to generate augmented trajectories. Specifically, we sample a scalar from the range $[0.3,1.7]$ and use it to scale each trajectory. To ensure the scaling does not introduce translation, we first center the trajectory by subtracting its mean, apply the scaling, and then restore its original position. To fully leverage the data augmentations, we train our models for $150$ epochs.

\subsubsection{Evaluation Metrics}

We evaluated our models using the standard evaluation protocol \cite{gupta2018social, salzmann2020trajectron, scholler2021flomo, socialvae2022, Shi_2023_ICCV, lin2024progressive}. We sampled $20$ trajectories from our model and reported the errors in meters using the following metrics:
\begin{itemize}[leftmargin=1em]
    \item \textbf{Minimum Average Displacement Error (minADE)} - Error of the sample with the smallest average L2 displacement error across all positions in the ground truth and forecasted trajectory.
    \item \textbf{Minimum Final Displacement Error (minFDE)} - Error of the sample with the smallest L2 displacement error between the last position in the ground truth and forecasted trajectory.
\end{itemize}
These evaluation metrics enable a direct comparison of forecasting accuracy with existing state-of-the-art models. Both minADE and minFDE are reported in meters.

\subsubsection{Ablations}

For all four model configurations, we trained and evaluated across all five folds of the ETH/UCY dataset, with the averaged results presented in Table \ref{tab:model_eth_ucy_performance}. Our findings indicate that neural differential equations are beneficial for both the causal encoder and the normalizing flow. However, the performance improvement when transitioning from a neural network to a neural differential equation is relatively small, suggesting that all configurations may be approaching the entropy floor of the dataset. Furthermore, while the joint formulation demonstrates a modest performance advantage over the marginal formulation on this dataset, the overall performance of the two approaches remains largely comparable.

\begin{table}[t]
\centering
\begin{tabular}{lccccc}
\toprule
\textbf{Model} & \textbf{minADE} & \textbf{minFDE} & \textbf{Parameters} & \textbf{\makecell{Training \\ Time (H)}} & \textbf{\makecell{Inference \\ Time (ms)}} \\
\midrule
S-GAN \cite{gupta2018social} & 0.48 & 0.96 & 85,499 & 1.09 & 13 \\
Trajectron++ \cite{salzmann2020trajectron} & 0.19 & 0.41 & 127,654 & 1.12 & 52 \\
FloMo \cite{scholler2021flomo} & 0.22 & 0.37 & 147,848 & 0.88 & 21 \\
S-VAE \cite{socialvae2022} & 0.21 & 0.33 & 2,144,002 & 2.78 & 25 \\
TUTR \cite{Shi_2023_ICCV} & 0.21 & 0.36 & 441,113 & 0.59 & \textbf{3} \\
PPT \cite{lin2024progressive} & 0.20 & \textbf{0.31} & 2,957,102 & 3.51 & 87 \\
\midrule
\textbf{GRU-DNF} & 0.21 & 0.39 & 287,160 & \textbf{0.58} & 18 \\
\textbf{GRU-CNF} & 0.20 & 0.39 & \textbf{53,620} & 4.87 & 137 \\
\textbf{CDE-DNF} & 0.19 & 0.39 & 293,624 & 6.27 & 73 \\
\textbf{CDE-CNF} & 0.19 & 0.38 & 60,084 & 11.22 & 201 \\
\textbf{CDE-CNF (Joint)} & \textbf{0.18} & 0.37 & 63,968 & 10.05 & 156 \\
\bottomrule
\end{tabular}
\caption{Average minADE and minFDE scores for various model configurations and baselines on the ETH/UCY datasets.}
\label{tab:model_eth_ucy_performance}
\end{table}

This increase in performance does not come without a cost. Because neural differential equations are evaluated by stepping through a numerical integrator, their training and inference times are significantly longer than neural networks. For example, the \textbf{CDE-CNF} model was approximately $19$ times slower than the \textbf{GRU-DNF} model during training time and roughly $11$ times slower during inference time. However, it is important to note that the numerical integrator employed in this study is an inefficient Python implementation. As a result, the observed discrepancy between training and inference times is likely larger than it would be with a more optimized implementation. Despite the increased training and inference times, neural differential equations utilize significantly less memory compared to traditional neural networks, approaching the efficiency of recurrent neural networks.

\subsubsection{Baseline Comparison}

Our best model (\textbf{CDE-CNF}), under both the joint and marginal formulations, performs better than all 6 baseline models on the minADE evaluation metric and is competitive on the minFDE metric.  The performance improvement demonstrated by \textbf{TrajFlow} over the baseline techniques on minADE is marginal and all methods have competitive performance on minFDE. This further supports the conclusion that models may be approaching the entropy floor of the dataset. Although our best model is more parameter-efficient than all existing baselines, it exhibits inferior performance in terms of training and inference time.

\subsection{Vehicle Experiment}

For the inD dataset we train and evaluate on a single recording session randomly splitting it into a $75
\%$ training set and a $25\%$ test set. We utilized a sequence length of $100$ for both the observed and unobserved trajectories and require all $200$ time steps to be present during both training and evaluation. To exclude outliers, such as parked vehicles, we ignore all agents with trajectories exceeding $1,000$ time steps.

We did not observe any benefit from providing rotational invariance, as in the ETH/UCY experiments. However, we found it necessary to normalize the data using min-max normalization:
\begin{equation}
    x' = \frac{x - x_{min}}{x_{max} - x_{min}}.
\end{equation}
Conversely, for the joint formulation, translational invariance again proved to be beneficial. Given the larger size of the inD dataset, we opted not to apply data augmentation and instead trained our models for 25 epochs.

\subsubsection{Evaluation Metrics}

We believe the standard minADE/minFDE metrics used with the ETH/UCY dataset are severely limited due to their inability to converge to a statistically meaningful value, instead diminishing toward zero as the sample size increases. Therefore, for the inD dataset, we sampled the learned distribution $1,000$ times for each test instance in the testing set and evaluated models with the following metrics:
\begin{itemize}[leftmargin=1em]
    \item \textbf{\textit{Root Mean Squared Error (RMSE)}} - The square root of the mean squared error between sampled points and corresponding ground truth trajectory
    \begin{equation}
        \text{RMSE} = \sqrt{\frac{1}{n} \sum_{i=1}^n (y_i - \hat{y}_i)^2}.
    \end{equation}
    \item \textbf{\textit{Continuous Ranked Probability Score (CRPS)}} - The difference between the cumulative distribution function (CDF) of the ground truth trajectory and the empirical CDF of the sampled points \cite{matheson1976scoring}
    \begin{equation}
        \text{CRPS} = \mathbb{E}[|X-y|] + \mathbb{E}[X] - 2 \mathbb{E}[X \cdot F(X)].
    \end{equation}
\end{itemize}
The RMSE metric (reported in meters) allows us to determine the performance of our models at predicting future agent locations even though they were not directly trained on this task. The CRPS metric allows us to evaluate the quality of the densities learned by our models.

\begin{figure}[t]
    \centering
    \includegraphics[width=\textwidth]{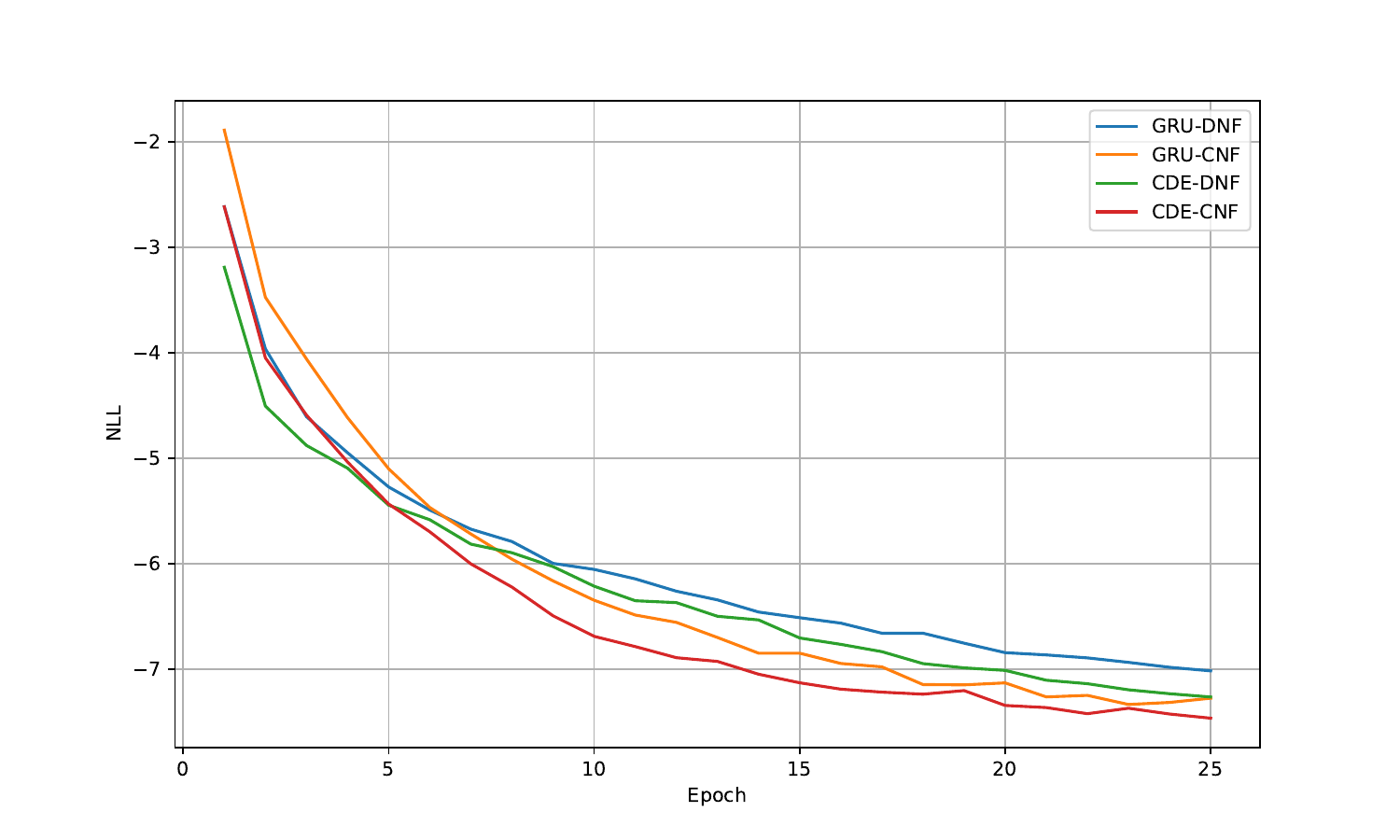}
    \caption{Negative log likelihood of our models when trained for 25 epochs on the inD dataset.}
    \label{fig:model-loss}
\end{figure}

\subsubsection{Ablations}

We trained and evaluated all four model configurations, with the results summarized in Table \ref{tab:model_ind_performance}. Consistent with the findings from the pedestrian experiment, neural differential equations offered performance advantages. As shown in Figure \ref{fig:model-loss}, replacing a neural network with a neural differential equation resulted in a predictable reduction in loss.

\begin{table}[!t]
\centering
\begin{tabular}{lccccc}
\toprule
\textbf{Model} & \textbf{RMSE} & \textbf{CRPS} & \textbf{Parameters} & \textbf{\makecell{Training \\ Time (H)}} & \textbf{\makecell{Inference \\ Time (ms)}} \\
\midrule
S-GAN \cite{gupta2018social} & 3.11 & 1.99 & 3,178,783 & 1.83 & 35 \\
Trajectron++ \cite{salzmann2020trajectron} & 1.72 & 0.72 & 3,286,342 & 0.59 & 33 \\
FloMo \cite{scholler2021flomo} & 2.39 & 0.56 & 12,723,928 & 0.36 & \textbf{20} \\
S-VAE \cite{socialvae2022} & 1.91 & 0.87 & 2,144,002 & 0.84 & 210 \\
TUTR \cite{Shi_2023_ICCV} & 1.93 & 0.43 & 1,918,665 & 0.15 & 3 \\
PPT \cite{lin2024progressive} & 1.16 & 0.54 & 3,001,134 & 0.15 & 160 \\
\midrule
\textbf{GRU-DNF} & 1.30 & 0.43 & 16,521,240 & \textbf{0.03} & 77 \\
\textbf{GRU-CNF} & 1.31 & 0.43 & \textbf{1,676,052} & 0.7 & 284 \\
\textbf{CDE-DNF} & 1.17 & 0.39 & 17,551,512 & 8.5 & 232 \\
\textbf{CDE-CNF} & \textbf{1.14} & \textbf{0.38} & 2,706,324 & 9.6 & 397 \\
\textbf{CDE-CNF (Joint)} & 1.47 & 0.45 & 2,957,248 & 6.04 & 141 \\
\bottomrule
\end{tabular}
\caption{RMSE and CRPS scores for various model configurations and baselines on the inD dataset.}
\label{tab:model_ind_performance}
\end{table}

The performance metrics highlight a substantial improvement when utilizing a neural controlled differential equation as a causal encoder, in contrast to the marginal performance gains observed with continuous normalizing flows. This result aligns with expectations, as neural controlled differential equations enable the continuous encoding of trajectory data, which is consistent with the inherently continuous nature of the data modality. Conversely, continuous normalizing flows primarily provide advantages for modeling multimodal or discontinuous densities—scenarios where affine transformations face challenges. However, the marginal formulation simplifies the learned density to a unimodal and continuous form, diminishing the potential benefits of continuous normalizing flows in this context. 

The joint formulation of \textbf{TrajFlow} exhibits inferior performance compared to all four configurations of the marginal formulation on the inD dataset, emphasizing the advantages of the marginal formulation in simplifying density representations—a key factor in enhancing performance for long time-horizon trajectory forecasting.

\subsubsection{Baseline Comparison}

The marginal formulation of the \textbf{CDE-CNF} configuration achieves superior perfromance on the RMSE and CRPS metric over all 6 of the baseline models. The most competitive baseline in terms of RMSE is PPT \cite{lin2024progressive}, whereas TUTR \cite{Shi_2023_ICCV} performs best with respect to CRPS. However, it is important to note that both of these models are deterministic and thus have limited utility in modeling the stochastic nature of dynamic motion. All four configurations of \textbf{TrajFlow}, as well as the joint formulation of the \textbf{CDE-CNF} configuration outperform all 4 baseline generative models (S-GAN \cite{gupta2018social}, Trajectron++ \cite{salzmann2020trajectron}, FloMo \cite{scholler2021flomo}, S-VAE \cite{socialvae2022}) on the MSE and CRPS metric. These results underscore the superior performance of \textbf{TrajFlow}. Moreover, the joint formulation of \textbf{TrajFlow} outperforms FloMo, demonstrating the effectiveness of continuous normalizing flows in capturing complex, multimodal, or discontinuous densities.

\section{Discussion}

Our experiments revealed that both the marginal formulation and neural differential equations positively influenced model performance. While the marginal and joint formulations produced comparable results on the ETH/UCY dataset, the marginal formulation demonstrated a moderate performance improvement on the inD dataset. This improvement is likely attributable to the simplified marginal density, which aids in long-horizon trajectory forecasting.

Likewise, we found that replacing discrete components with neural differential equations generally improved model performance and reduced model loss. On the ETH/UCY dataset, the performance improvement was minimal, likely indicating that the entropy floor of the dataset had been reached. In contrast, a more pronounced performance gain was observed on the inD dataset, with neural controlled differential equations yielding a significantly greater impact than continuous normalizing flows. This outcome aligns with expectations, as neural controlled differential equations enable the model to capture the full range of information present in continuous trajectory data. In comparison, continuous normalizing flows are more effective at learning multi-modal or discontinuous densities. However, the marginal formulation constrains the learned density to a unimodal and continuous form, limiting the benefits of continuous normalizing flows in this context.

Although neural differential equations offer clear performance benefits, they come with significant computational overhead as they require numerical integration for evaluation. As a result, the continuous formulation of \textbf{TrajFlow} is best suited for batch processing or environments with limited memory resources, while the discrete formulation may be better suited for real-time systems.

The advantages of marginal densities extend beyond mere quantitative improvements in model performance. As illustrated in Figure \ref{fig:spatial-density-evolution}, the marginal formulation captures the evolution of occupancy density over time. This capability enables the computation of motion trajectories by sampling positions at each time step, as well as the generation of occupancy grids, thereby supporting the two most common representations in motion forecasting \cite{mahjourian2022occupancy}. What's more, because the marginal formulation incorporates time $s$ as a parameter in its normalizing flow, it enables fully continuous sampling of future locations.

\begin{figure}[!t]
    \centering
    \includegraphics[width=1.0\linewidth]{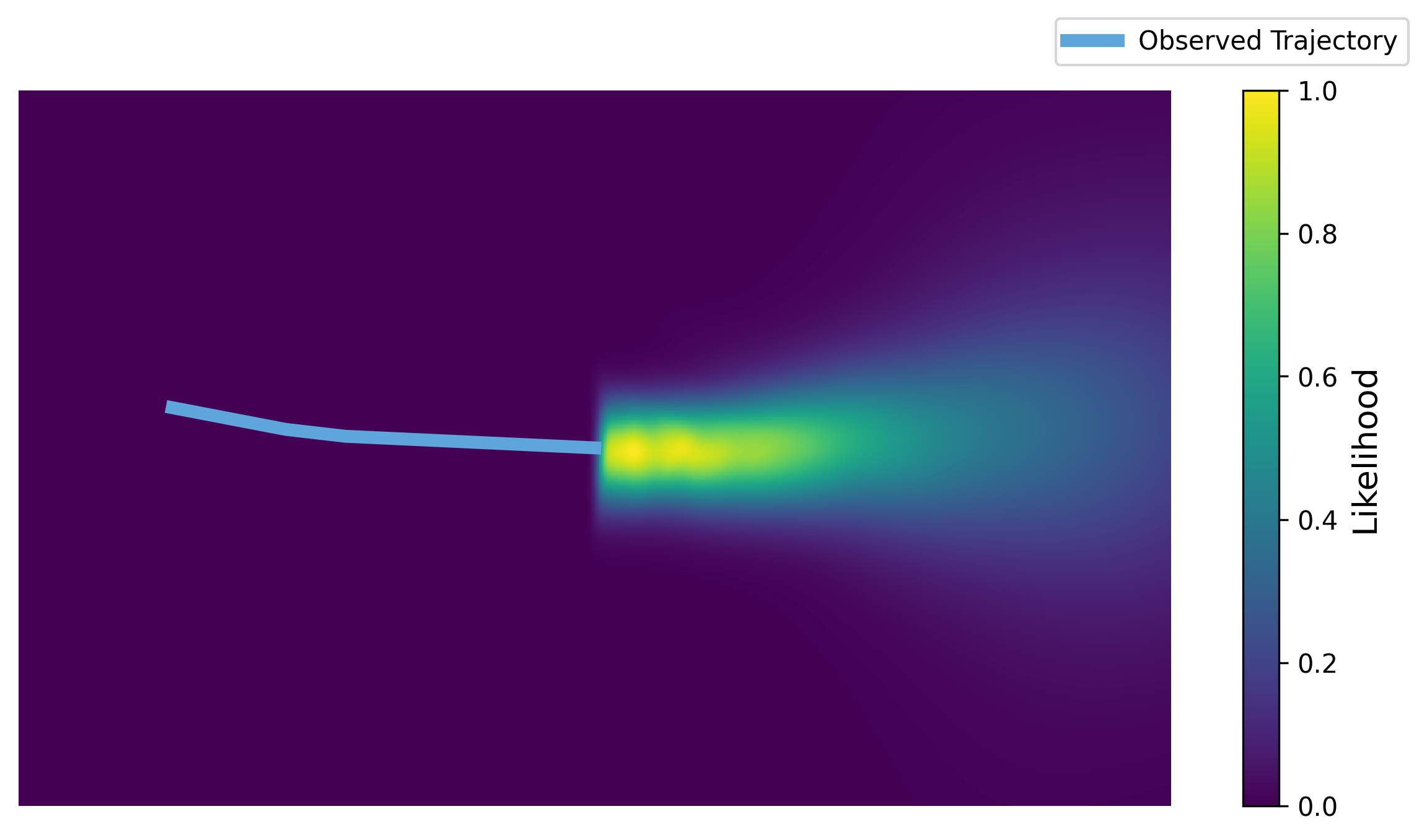}
    \caption{Occupancy grid produced by additive fusion of the estimated occupancy densities at $10$ times sampling frequency.}
    \label{fig:occupancy-map}
\end{figure}

The ability to compute occupancy grids from marginal densities further underscores their practical utility in applications such as motion planning and black-spot identification. Occupancy grids provide agents with a probabilistic map of spatial locations, enabling informed trajectory planning by highlighting areas of likely occupancy. One effective method for generating occupancy grids involves applying additive fusion to the future occupancy densities predicted by \textbf{TrajFlow}. Additive fusion entails summing all occupancy densities and normalizing the result by the maximum cell value, ensuring a cohesive probability distribution. Sampling at a frequency higher than the training frequency produces smoother occupancy grids, enhancing their utility in complex scenarios. For instance, Figure \ref{fig:occupancy-map} illustrates an occupancy grid for a single agent, generated by fusing occupancy densities sampled at 10 times the training frequency. In the limit, our additive fusion strategy can be described by the following equation:
\begin{equation}
    \frac{\int_{0}^{S} p(u^s_i \mid O_i, F_i) \, ds}{\max\limits_{u_i, s} p(u^s_i \mid O_i, F_i)}.
\end{equation}

\begin{figure}[!t]
    \centering
    \includegraphics[width=1.0\linewidth]{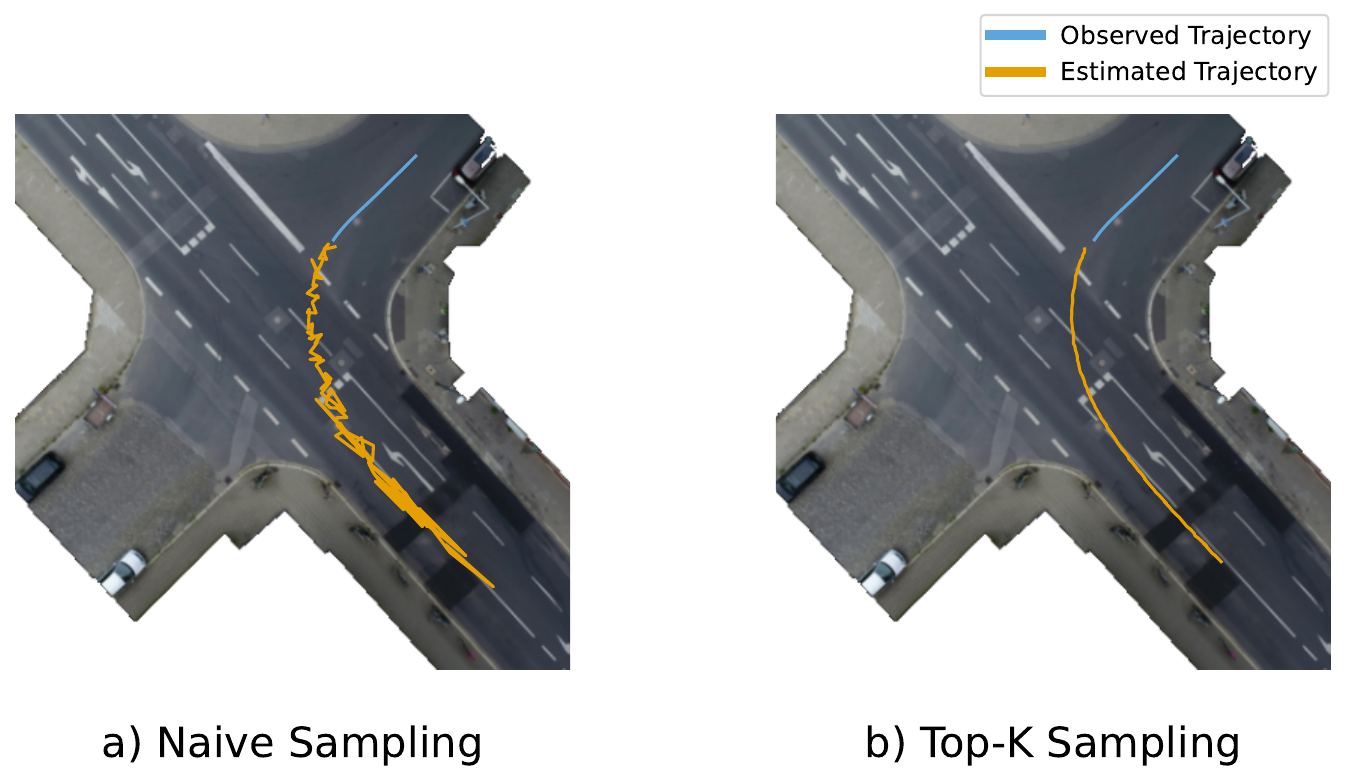}
    \caption{Comparison of sampling strategies for estimating trajectories with our model. Top-k sampling significantly reduces the noise in sampled trajectories.}
    \label{fig:trajectory-sampling}
\end{figure}

%
One limitation of modeling the marginal density is that it inherently reduces diversity among the sampled motion trajectories and introduces noise, as the relationships between predicted locations are no longer explicitly modeled. This reduction in diversity can impact the model's ability to represent the full range of plausible trajectories. However, this issue can be mitigated by employing a top-$k$ sampling strategy, which significantly reduces the noise in the sampled trajectories, as illustrated in Figure \ref{fig:trajectory-sampling}. In this approach, instead of sampling a single point at each time step, the model samples $k$ points and selects the most likely one based on the marginal density, thereby improving trajectory coherence. Despite this improvement, the top-$k$ approach has its drawbacks, as it can lead to a mode collapse where the model predominantly predicts the most likely trajectory, potentially overlooking less probable but valid alternatives.

\section{Conclusion}

In this work, we introduce \textbf{TrajFlow}, a probabilistic framework for modeling occupancy densities. It estimates the likelihood that an agent will be at a specific location at a given time, conditioned on a sequence of observed historical locations. We provided a discrete and continuous implementation of this framework. The continuous implementation was noticeably more performant, but came at a computational cost.

To evaluate our framework, we conducted extensive ablations and experiments, demonstrating that it achieves state-of-the-art performance on common trajectory forecasting benchmarks. We highlighted the advantages of a marginal formulation, showcasing its ability to estimate occupancy densities over continuous time. This capability allows our framework to produce both smooth motion trajectories and occupancy grids, thereby supporting two of the most common representations in motion forecasting. Since \textbf{TrajFlow} does not model the relationship between predicted locations, the diversity in the motion trajectories sampled may be limited. Future work is needed to explore the real-world applicability and usefulness of the predicted motion trajectories and occupancy grids in practical applications such as motion planning and black-spot identification.

\section{Declaration of generative AI and AI-assisted technologies in the writing process}
During the preparation of this work, the authors utilized AI-powered natural language processing tools such as Grammarly and LLMs strictly for phrasing assistance and editing. After using these tools, the authors thoroughly reviewed and revised the content as necessary and take full responsibility for the final revision of the published article. 

\section{Author Contribution Statement}
The authors confirm the contributions to the paper are as follows: study conception and design: Mitch Kosieradzki, Seongjin Choi; data collection: Mitch Kosieradzki; data visualization: Mitch Kosieradzki; analysis and interpretation of the results: Mitch Kosieradzki; draft manuscript preparation: Mitch Kosieradzki, Seongjin Choi. All authors have reviewed the results and approve of the final version of the manuscript.

\bibliographystyle{trb}
\bibliography{trb_template}
\end{document}